\documentclass[11pt,a4paper]{article}
\usepackage[hyperref]{acl2021}
\usepackage{times}
\usepackage{latexsym}

\usepackage{microtype}

\aclfinalcopy

\usepackage{algpseudocode}
\usepackage{algorithm}
\usepackage{multirow}
\usepackage{graphicx}

\title{RRULES: An improvement of the RULES rule-based classifier}

\author{Rafel Palliser-Sans \\
  Facultat d'Informàtica de Barcelona (FIB) \\
  Universitat Politècnica de Catalunya (UPC - BarcelonaTECH) \\
  Barcelona, Spain \\
  \texttt{rafel.palliser@estudiantat.upc.edu} \\}

\date{}

\begin{document}
\maketitle
\begin{abstract}
RRULES is presented as an improvement and optimization over RULES, a simple inductive learning algorithm for extracting IF-THEN rules from a set of training examples. RRULES optimizes the algorithm by implementing a more effective mechanism to detect irrelevant rules, at the same time that checks the stopping conditions more often. This results in a more compact rule set containing more general rules which prevent overfitting the training set and obtain a higher test accuracy. Moreover, the results show that RRULES outperforms the original algorithm by reducing the coverage rate up to a factor of 7 while running twice or three times faster consistently over several datasets.
\end{abstract}

\section{Introduction}
\textbf{Rule-based classifiers} are Machine Learning algorithms that induce a collection of \textbf{IF-THEN rules} from a set of training examples to then apply them to classify new cases into one of the possible classes. They are one of the most classic and basic classifiers, yet effective and widely used in expert systems nowadays.

This work analyzes the RULES \citep{rules} classifier and proposes an improvement. Our main contribution is the presentation of \textbf{RRULES}, an optimized algorithm that outperforms the original one by obtaining \textbf{better performance} and \textbf{working faster}.

The source code for both the original RULES algorithm and the RRULES optimization is available at \url{https://github.com/rafelps/RRULES-rule-based-classifier}.

\section{Related Work}

Rule-based classifiers can be seen as a generalization of Decision Tree classifiers---like the well-known ID3 \citep{id3} or C4.5 \citep{c45}--- considering that the obtained rules do not require to be represented in the form of a tree, thus being more flexible.

Regarding the induced rules, these can have different structures, but this work focuses on methods that use conjunctive antecedents: ``\textbf{IF} condition\textsubscript{1} \textbf{AND} condition\textsubscript{2} \textbf{AND} ... \textbf{AND} condition\textsubscript{n} \textbf{THEN} conclusion''. Each individual condition, also called selector, is formed by a pair Attribute-Value, referring to a specific attribute of the data, and the value that it takes.

One of the first rule-based classifiers that appeared was PRISM \citep{prism}, which selects all the examples belonging to a certain class to obtain rules that can classify them. PRISM starts generating general rules, and adds selectors or individual conditions until the obtained rule is ``perfect'', which in their algorithm means that the rule has a 100\% precision: the rule only affects to instances of the same class. PRISM is a non-ordered algorithm---the order in which rules are induced is not relevant for prediction---, non-incremental---it requires all the data when starting the induction---, and goes from general to specific rules.

CN2, presented two years later by \citet{cn2}, introduces the concept of ``complex'' as a combination of selectors, or antecedent of a rule. In their work, the authors implement an heuristic based on entropy measures to perform a k-beam search of the ``best complex''. Whenever the ``best complex'' is found, the consequent of the rule is formed by the mode class of the instances that it covers, therefore not enforcing a 100\% precision. However, as a trade-off, CN2 obtains more general rules that often cover more instances and help generalizing over unseen cases. It is a non-incremental but ordered algorithm, which implies that when predicting the algorithm applies the rules in the order they were induced.

Rise \citep{rise} is another rule-based classifier that, opposite to the previously mentioned, starts building specific rules. In fact, it starts by creating a rule for each single example (including all its attributes in the antecedent) and iteratively merges the two of them that produce the highest precision in the resulting set of rules. Apart from being non-ordered, Rise is an incremental algorithm, which means that new examples can be introduced at any time. In that case, these are converted into specific rules and the algorithm resumes using the previously obtained rule set and the new examples.

\subsection{RULES}
Rules Extraction System (RULES) \citep{rules} is a pretty intuitive algorithm to induce rules from a set of training instances. It goes from general to specific and enforces a 100\% precision in the generated rules (unless there are inconsistencies in the data and two instances with the same attributes belong to different classes).

To do so, RULES starts checking if there is any rule of a single selector (pair Attribute-Value) that has perfect precision. If not, it checks combinations of more selectors, increasing one at a time. The complete algorithm can be seen in Algorithm \hyperref[alg:rules]{\ref{alg:rules} (RULES)}\footnote{Lines \ref{line:19} to \ref{line:23} of Algorithm \ref{alg:rules} are not present in the pseudocode of the original paper. These are added to ensure a best-effort classification when there are inconsistencies in the data. Moreover, a default rule could be created to classify test examples that do not match any induced rule into the most frequent class in the training data.}. The concept ``Irrelevant Condition'', used in the algorithm, is defined by the authors as a condition whose selectors are partially included in a already-created rule. Note that if that is the case, we already have a more general rule that classifies the instances that match the more specific antecedent that was being created. For example, ``IF Attribute\textsubscript{1} is Value\textsubscript{11} AND Attribute\textsubscript{2} is Value\textsubscript{21}'' is an irrelevant condition if we already have a rule like ``IF Attribute\textsubscript{1} is Value\textsubscript{11} THEN Class is C\textsubscript{1}'', because the latter already classifies examples that would match the former condition.

\begin{algorithm*}[ht]
\begin{algorithmic}[1]
    \Require $T$ \Comment{T is the set of training examples}
    \State $N \gets T$ \Comment{$N$ is the set of non-classified examples}
    \State $Rules \gets \emptyset$ \Comment{$Rules$ is the set of rules}
    \For{$n_c$ in (1, ..., $n_a$)} \Comment{$n_a$ is the number of attributes for each element}
        \If{$|N| = 0$} \Comment{Everything classified} \label{line:stop}
            \State \Return $Rules$
        \EndIf
        \State Obtain all the selectors Attribute-Value present in $N$ and generate all possible conditions $Cond$ as combinations of $n_c$ selectors.
        \ForAll{$Cond$}
            \State $M \gets$ Instances in $T$ matching $Cond$ as antecedent
            \If{$|M| = 0$} \Comment{No element in the dataset meets the condition}
                \State Discard condition and move to following one
            \EndIf
            \If{All instances $m_i \in M$ belong to the same class $C$}
                \If{$Cond$ is not an Irrelevant Condition}
                    \State Create rule $R: Cond \rightarrow C$
                    \State $Rules \gets Rules + R$
                    \State $N \gets N - \{M\}$
                \EndIf
            \ElsIf{$n_c = n_a$} \Comment{Inconsistency: Same attributes, different class}\label{line:19}
                \State Find most probable class $C^*$ among $M$
                \State Create rule $R: Cond \rightarrow C^*$
                \State $Rules \gets Rules + R$
                \State $N \gets N - \{M\}$\label{line:23}
            \EndIf
        \EndFor
    \EndFor
    \State \Return $Rules$
\end{algorithmic}
\caption{RULES \cite{rules}}
\label{alg:rules}
\end{algorithm*}

\section{RRULES}
RRULES proposes an optimization over RULES \citep{rules} focusing in two key points: The mechanism to \textbf{detect irrelevant rules}, and the \textbf{stopping condition}.
\subsection{Irrelevant Rules}

Imagine there is a dataset like the one in Table~\ref{tab:dataset_example}. If RULES is applied, in the first iteration---considering only one selector in the antecedent---the rule ``IF $A$ is $A_1$ THEN Class is 0'' would be created. Following, examples 1 and 2 would be removed from the set of non-classified examples. There is not any other selector that can classify all matched instances with 100\% precision, so the algorithm moves to the second iteration. Now, RULES considers all combinations of length 2 of selectors present in the examples 3, 4 and 5. These are $A=A_2$, $B=B_1$, $B=B_2$, $C=C_1$, and $C=C_2$. Note that a one of the possible combinations is ($B=B_1$, $C=C_1$), which classifies all matched instances (examples 1 and 2) with 100\% precision, so the rule ``IF $B$ is $B_1$ AND $C$ is $C_1$ THEN Class is 0'' would also be created\footnote{This rule is not considered irrelevant because it uses different selectors than the already-created one.}.

Note that these two rules \textbf{classify the exact same instances}, and the second one---less general--- do not help classifying any of the non-classified instances. It is true that the two rules give different perspectives but we might be losing generality, and in the case of highly correlated attributes one of them would not be relevant at all.

For this reason, our first optimization is to change the ``Check Irrelevant Condition'' step of Algorithm \hyperref[alg:rules]{\ref{alg:rules} (RULES)} into checking if the \textbf{specific combination of selectors} that we are testing is present in the non-classified instances. It is important to remark that RULES obtains all selectors present in the non-classified instances but then generates all possible combinations, even those that are not present in the examples not yet classified.

This change can be seen in lines \ref{lines:irrelevant1} and \ref{lines:irrelevant2} of Algorithm \hyperref[alg:rrules]{\ref{alg:rrules} (RRULES)}.

\begin{table}
\centering
\begin{tabular}{ccccc}
\hline
\multirow{2}{*}{Example} & \multicolumn{3}{c}{Attributes} & \multirow{2}{*}{Class} \\
 & $A$ & $B$ & $C$ &  \\ \hline\hline
1 & $A_1$ & $B_1$ & $C_1$ & 0 \\ \hline
2 & $A_1$ & $B_1$ & $C_1$ & 0 \\ \hline
3 & $A_2$ & $B_1$ & $C_2$ & 1 \\ \hline
4 & $A_2$ & $B_2$ & $C_1$ & 2 \\ \hline
5 & $A_2$ & $B_2$ & $C_2$ & 3 \\ \hline

\end{tabular}
\caption{Dataset Example.}
\label{tab:dataset_example}
\end{table}

\subsection{Stopping Condition}
The original algorithm stops whenever the set of non-classified instances is empty, and checks it \textbf{between each outer iteration}, as seen in line \ref{line:stop} of Algorithm \hyperref[alg:rules]{\ref{alg:rules} (RULES)}.

This procedure has a weak point: if the iteration starts with a single non-classified instance (or a low number of them), the algorithm will perform \textbf{all the inner iterations} (checking every condition) when it could stop earlier saving computation. Moreover, with the original implementation of ``Check Irrelevant Condition'', these iterations could generate \textbf{several unnecessary rules}.

In our implementation we check for the stopping condition just after creating each rule, which lets us be more efficient. This can be seen in lines \ref{lines:stop1} and \ref{lines:stop2} of Algorithm \hyperref[alg:rrules]{\ref{alg:rrules} (RRULES)}.

\begin{algorithm*}[ht]
\begin{algorithmic}[1]
    \Require $T$ \Comment{T is the set of training examples}
    \State $N \gets T$ \Comment{$N$ is the set of non-classified examples}
    \State $Rules \gets \emptyset$ \Comment{$Rules$ is the set of rules}
    \For{$n_c$ in (1, ..., $n_a$)} \Comment{$n_a$ is the number of attributes for each element}
        \State Obtain all the selectors Attribute-Value present in $N$ and generate all possible conditions $Cond$ as combinations of $n_c$ selectors.
        \ForAll{$Cond$}
            \State $M_T \gets$ Instances in $T$ matching $Cond$ as antecedent
            \State $M_N \gets$ Instances in $N$ matching $Cond$ as antecedent \label{lines:irrelevant1}
            \If{$|M_T| = 0$} \Comment{No element in the dataset meets the condition}
                \State Discard condition and move to following one
            \EndIf
            \If{$|M_N| = 0$} \Comment{Irrelevant condition} \label{lines:irrelevant2}
                \State Discard condition and move to following one
            \EndIf
            \If{All instances $m_i \in M_T$ belong to the same class $C$}
                \State Create rule $R: Cond \rightarrow C$
                \State $Rules \gets Rules + R$
                \State $N \gets N - \{M_N\}$
                \If{$|N| = 0$} \Comment{Everything classified} \label{lines:stop1}
                    \State \Return $Rules$
                \EndIf
            \ElsIf{$n_c = n_a$} \Comment{Inconsistency: Same attributes, different class}
                \State Find most probable class $C^*$ among $M_T$
                \State Create rule $R: Cond \rightarrow C^*$
                \State $Rules \gets Rules + R$
                \State $N \gets N - \{M_N\}$
                \If{$|N| = 0$} \Comment{Everything classified} \label{lines:stop2}
                    \State \Return $Rules$
                \EndIf
            \EndIf
        \EndFor
    \EndFor
    \State \Return $Rules$
\end{algorithmic}
\caption{RRULES}
\label{alg:rrules}
\end{algorithm*}

\section{Experiments and Results}
\subsection{Data}
In the following experiments we use \textbf{two groups} of datasets. The \textbf{first} five sets---Season, Human, Golf, Cheese and Flower---are directly extracted from the \textbf{original publication} of RULES \citep{rules} and contain few examples (up to 29). These are used in a \textbf{train only} basis to assess how the algorithm works and which rules it induces.

The \textbf{second} group is formed by Breast-Cancer, Tic-Tac-Toe, Mushroom, and Iris. These four sets from the UCI Machine Learning Repository \citep{uci} contain up to 8000 instances and significantly more attributes. For these experiments, the data has been split 80\%-20\% into train and test, and for the Iris dataset we have discretized its attributes into 7 equal bins.

\begin{table}
\centering
\begin{tabular}{lcccc}
\hline
Dataset &
\rotatebox{90}{N Examples  } &
\rotatebox{90}{N Attributes  } &
\rotatebox{90}{N Selectors} &
\rotatebox{90}{N Classes}\\ \hline\hline
Season & 11 & 3 & 9 & 4 \\ \hline
Human & 8 & 3 & 7 & 2 \\ \hline
Golf & 14 & 4 & 10 & 2 \\ \hline
Cheese & 29 & 4 & 11 & 13 \\ \hline
Flower & 29 & 5 & 27 & 25 \\ \hline\hline
Breast-Cancer & 286 & 9 & 51 & 2 \\ \hline
Tic-Tac-Toe & 958 & 9 & 27 & 2 \\ \hline
Mushroom & 8124 & 22 & 126 & 2 \\ \hline
Iris & 150 & 4 & 28 & 3 \\ \hline

\end{tabular}
\caption{Datasets.}
\label{tab:datasets}
\end{table}

\subsection{Metrics}
To assess the performance of our algorithm we \textbf{compare its results to the original RULES} method by using several metrics. \textbf{On the one hand} we are going to compare the set of induced rules by means of its \textbf{size}, the mean \textbf{precision}, the overall \textbf{coverage} and the \textbf{time} required for the algorithm to induce it.

The mean precision is computed as the mean of per-rule precisions, calculated as the ratio between the number of instances correctly classified by a rule and the number of examples that match the antecedent or conditions. It measures the ability of that rule to correctly classify the instances that it covers. Higher is better, highest is 100\%.

The overall coverage is computed as the sum of per-rule coverages, each measured as the ratio between the number of instances that the rule covers (matching the antecedent) and the number of instances in the dataset. While large per-rule coverages are preferable because they indicate generality, the overall coverage should be close to 100\%, because larger ones imply that there are unnecessary or irrelevant rules. As both RULES and RRULES keep adding rules until every instance is classified, the lowest coverage for the training set is 100\%. Lower is better.

\textbf{On the other hand}, for the group of UCI datasets in which we can create a test split, we are also going to use the \textbf{test accuracy} as a metric to study the performance of these methods to correctly predict the class of unseen cases. Higher is better, highest is 100\%.

\subsection{Results}
Table \ref{tab:results} shows the comparison for the results obtained by the RULES \citep{rules} and our RRULES algorithms. It can be seen that in every dataset our algorithm obtains a \textbf{more compact rule set}. In the case of Breast-Cancer, RRULES generates almost 8 less times the amount of rules induced by RULES, which generates an average of 3 rules per example in the dataset.

Regarding \textbf{precision}, whenever the algorithm cannot obtain a 100\% due to inconsistencies in the data (or discretization not being perfect), our approach obtains \textbf{slightly better results} than the original paper.

Coverage metrics show a similar information than the size of the obtained rule set. In all cases our algorithm reaches a \textbf{coverage closer to 100\%}, being significantly better in Breast-Cancer and Mushroom sets. In the former, each instance is covered by an average of 11 rules, while in the latter by 18. Many of them are probably unnecessary or irrelevant.

Despite generating less rules, the \textbf{test accuracy} shows that our algorithm is at least as accurate as the original one, even outperforming it in some cases. Therefore, by creating fewer rules we gain generalization power to predict unseen examples.

Finally, our algorithm needs \textbf{half or even three times less time} to induce the rule set in some datasets thanks to our optimizations and checking the stopping condition more often.

\begin{table*}
\centering
\begin{tabular}{clccccc}
\hline
Dataset & Algorithm & N Rules & Train Prec. & Train Cov. & Test Acc. & Ind. Time \\\hline\hline

 & RULES & 7 & 100\% & 127\% & - & 0s \\ \cline{2-7} 
\multirow{-2}{*}{Season} & RRULES & 6 & 100\% & 118\% & - & 0s \\ \hline\hline
 & RULES & 4 & 100\% & 112\% & - & 0s \\ \cline{2-7} 
\multirow{-2}{*}{Human} & RRULES & 4 & 100\% & 112\% & - & 0s \\ \hline\hline
 & RULES & 13 & 100\% & 221\% & - & 0s \\ \cline{2-7} 
\multirow{-2}{*}{Golf} & RRULES & 7 & 100\% & 121\% & - & 0s \\ \hline\hline
 & RULES & 24 & 100\% & 103\% & - & 0s \\ \cline{2-7} 
\multirow{-2}{*}{Cheese} & RRULES & 24 & 100\% & 103\% & - & 0s \\ \hline\hline
 & RULES & 56 & 100\% & 200\% & - & 0s \\ \cline{2-7} 
\multirow{-2}{*}{Flower} & RRULES & 27 & 100\% & 100\% & - & 0s \\ \hline\hline
 & RULES & 929 & 99\% & 1169\% & 68.97\% & 1.36s \\ \cline{2-7} 
\multirow{-2}{*}{Breast-Cancer} & RRULES & 118 & 98\% & 198\% & 70.69\% & 0.26s \\ \hline\hline
 & RULES & 165 & 100\% & 231\% & 95.31\% & 0.22s \\ \cline{2-7} 
\multirow{-2}{*}{Tic-Tac-toe} & RRULES & 73 & 100\% & 164\% & 96.88\% & 0.19s \\ \hline\hline
 & RULES & 531 & 100\% & 1879\% & 99.88\% & 0.43s \\ \cline{2-7} 
\multirow{-2}{*}{Mushroom} & RRULES & 78 & 100\% & 253\% & 99.88\% & 0.22s \\ \hline\hline
 & RULES & 39 & 98\% & 227\% & 100\% & 0.04s \\ \cline{2-7} 
\multirow{-2}{*}{Iris} & RRULES & 22 & 96\% & 157\% & 100\% & 0.02s \\ \hline
\end{tabular}
\caption{Results for RULES\citep{rules} and RRULES in multiple datasets.}
\label{tab:results}
\end{table*}

\section{Conclusions}
In this work we have presented RRULES, an optimization over RULES \citep{rules} which is able to outperform the original algorithm by generating a more compact set of rules. These fewer rules are able to improve the generalization and prediction accuracy at the same time that the induction phase works faster than the original algorithm.

We have proven that creating too many specific rules can lead to a slight overfit to the training set while generating a reduced set of more-general rules serves as regularization, at the same time that requires less computation.

\bibliographystyle{acl_natbib}
\bibliography{mybib}


\end{document}